\pdfoutput=1
\documentclass[11pt]{article}

\usepackage[]{acl}

\usepackage{times}
\usepackage{latexsym}

\usepackage[T1]{fontenc}
\usepackage[utf8]{inputenc}

\usepackage{inconsolata}
\usepackage{times}
\usepackage{listings}
\usepackage{color}
\usepackage{latexsym}

\usepackage{xcolor}
\colorlet{punct}{red!60!black}
\definecolor{background}{HTML}{EEEEEE}
\definecolor{delim}{RGB}{20,105,176}
\colorlet{numb}{magenta!60!black}

\lstdefinelanguage{json}{
    basicstyle=\normalfont\ttfamily,
    numbers=left,
    numberstyle=\scriptsize,
    stepnumber=1,
    numbersep=8pt,
    showstringspaces=false,
    breaklines=true,
    frame=lines,
    backgroundcolor=\color{background},
    literate=
     *{0}{{{\color{numb}0}}}{1}
      {1}{{{\color{numb}1}}}{1}
      {2}{{{\color{numb}2}}}{1}
      {3}{{{\color{numb}3}}}{1}
      {4}{{{\color{numb}4}}}{1}
      {5}{{{\color{numb}5}}}{1}
      {6}{{{\color{numb}6}}}{1}
      {7}{{{\color{numb}7}}}{1}
      {8}{{{\color{numb}8}}}{1}
      {9}{{{\color{numb}9}}}{1}
      {:}{{{\color{punct}{:}}}}{1}
      {,}{{{\color{punct}{,}}}}{1}
      {\{}{{{\color{delim}{\{}}}}{1}
      {\}}{{{\color{delim}{\}}}}}{1}
      {[}{{{\color{delim}{[}}}}{1}
      {]}{{{\color{delim}{]}}}}{1},
}

\usepackage{adjustbox}
\usepackage{array}
\usepackage{multirow}

\newcolumntype{R}[2]{%
     >{\adjustbox{angle=#1,lap=\width-(#2)}\bgroup}%
     l%
     <{\egroup}%
 } 
\newcommand*\rotr{\multicolumn{1}{R{45}{1em}}}% no optional argument here
\newcommand*\rotrow{\rotatebox[origin=c]{90}}% no optional argument here
\usepackage{graphicx}
\usepackage{multirow}
\usepackage{tabularx}
\usepackage{booktabs}
\usepackage{xspace}
\usepackage{amsmath,amsfonts,amssymb}
\usepackage{float}
\usepackage{stfloats}
\usepackage{slashed}
\usepackage{amsbsy}
\usepackage{amstext}
\usepackage{microtype}
\usepackage{pgfplots}
\pgfplotsset{width=1.0\columnwidth}
\usepackage[multiple]{footmisc}
\usepackage{enumitem}
\usepackage{tabularx}
\usepackage{bbding}

\floatstyle{ruled}
\newfloat{algorithm}{tbp}{loa}
\providecommand{\algorithmname}{Algorithm}
\floatname{algorithm}{\protect\algorithmname}

\title{Dataset and Baseline System for Multi-lingual Extraction and Normalization of Temporal and Numerical Expressions}

\author{Sanxing Chen$^{*}$ \\
  Duke University \\
  \texttt{sanxing.chen@duke.edu} \\\And
  Yongqiang Chen$^{*}$ \\
  The Chinese University\\ of Hong Kong \\
  \texttt{yqchen@cse.cuhk.edu.hk} \\\And
  Börje F. Karlsson \\
  Microsoft Research Asia \\
  \texttt{borjekar@microsoft.com} \\}

\begin{document}
\maketitle

\begin{abstract}

Temporal and numerical expression understanding is of great importance in many downstream Natural Language Processing (NLP) and Information Retrieval (IR) tasks. However, much previous work covers only a few sub-types and focuses only on entity extraction, which severely limits the usability of identified mentions. In order for such entities to be useful in downstream scenarios, coverage and granularity of sub-types are important; and, even more so, providing resolution into concrete values that can be manipulated.
Furthermore, most previous work addresses only a handful of languages.
% And more importantly, lack good evaluation datasets.
Here we describe a multi-lingual evaluation dataset - NTX - covering diverse temporal and numerical expressions across 14 languages and covering extraction, normalization, and resolution.
Along with the dataset we provide a robust rule-based system as a strong baseline for comparisons against other models to be evaluated in this dataset. Data and code
%, and pre-built binaries 
will be publicly available at \url{https://aka.ms/NTX}.

\let\thefootnote\relax\footnotetext{*The work described in this technical report was performed during the authors' internships at Microsoft Research Asia.}

\end{abstract}

\section{Introduction}
\label{sec:introduction}

Entity recognition (or entity extraction) is a key component in many NLP pipelines and important for various downstream tasks. However, most ER works focus only on \emph{named} entities with types like Person, Organization, etc.; treating other potentially important terms (like datetime mentions or numerals) as only literals. 

This is problematic as such entities play important roles in information retrieval, relationship extraction, conversational language understanding, task completion, knowledge base construction, and beyond~\cite{10.1145/1328964.1328968, 10.1145/3290605.3300439, gesese2021}.

Furthermore, even when covering numerical and temporal entities, datasets treat them in a too unbalanced or coarsely way - e.g., OntoNotes 5~\cite{ontonotes5} has five categories for numerical entities, but only two for dates and times. On the numerical types side, such categorization ignores that numeric literals often denote specific units of measurements or types in context. E.g., "She is eight", clearly implies that "eight" is not just a cardinal number, but a description of \emph{age}.

Meanwhile, while there has been an increasing interest in temporal entities~\cite{uzzaman-etal-2013-semeval}, most efforts utilize complex annotations tagging both datetime expressions per se and also temporal relationships. Such coupled complex annotation schemas both are too intricate to be used effectively by downstream users and do not cover many mention forms necessary in practice.

Due to the complexity of existing annotation schemas and the cost of data acquisition and annotation, quality assets even for the entity types covered only exist in English and sparsely in a handful of languages.

Another parallel issue affecting numerical and temporal entities or expressions is that for them to be of actual use for downstream tasks, mention detection is not enough. Such mentions need to be normalized (turned into a canonical form that standardizes interpretation) and further resolved into concrete values for usage (sometimes necessitating extra context for resolution). E.g. "from February to the end of 2012" -> (XXXX-02,2012-EOY,P11M) -> start: 2012-02-01, end: 2012-12-31. The latter allows the conversion into specific datetime object instances that can be consumed in conventional programs.

Some of the previous efforts in creating schemas for temporal mentions, e.g., \citet{uzzaman-etal-2013-semeval}, also cover normalization/resolution. But they suffer from the same complexity and coverage problems mentioned, or also from mixing normalization and resolution. We emphasize the requirement to separate the two as resolution will in many cases require extra context and identifying this context is error prone. So any system addressing temporal entities should allow re-resolution from the normalized form. For example, "now" would have normalized form "PRESENT\_REF". But to turn it into a concrete resolution, an anchor reference datetime such as "2022-07-01" is needed. As such references can be incorrectly inferred, systems should allow re-resolution. A normalized form like "PRESENT\_REF" allows it, while the typical annotation in TIMEML would mix the two and directly consider the time expression as "2022-07-01" and re-resolution is not possible without re-processing the original text.

Due to the mentioned limitation of previous efforts and lack of existing datasets, here we propose a multi-lingual benchmark dataset that covers numerical and temporal entities and expressions at the extraction, normalization, and resolution levels. 

This dataset has been manually annotated in the context of a multi-year effort over real-world use in commercial applications\footnote{The Recognizers-Text project, available at \url{https://github.com/microsoft/Recognizers-Text}.}. The proposed schema focuses on real-world entity coverage and ease of use for downstream applications and includes: 8 numerical sub-types (cardinal, ordinal, percentage, numerical range, age, currency, dimension, temperature) and 10 temporal sub-types (date, time, datetime, date range, time range, datetime range, holiday, duration, timezone, and recurring set). 

The NTX dataset covers 14 languages\footnote{See detailed list of languages in Section \ref{sec:dataset details}.}, which belong to a diverse set of language families---Indo-European, Sino-Tibetan, Japonic, Turkic, Semitic, and Koreanic.

Furthermore, to both serve as a baseline for performance comparisons over this dataset and to help generate training data for new models following its schema, we reference a high quality rule-based system~\cite{recognizers-text} that achieves strong results and has been hardened through real-world commercial use.

\section{Related Work}
\label{sec:related work}

%As the main contribution of this work is the new dataset that addresses limitations of available data...

\subsection{Numerical Entities}
\label{subsec:numerical Entities}

Already in the nineties venues like MUC attempted to standardize the evaluation of information extraction (IE) tasks, including numerical expression extraction (and in six languages) \cite{10.3115/974557.974585}. Perhaps due to high coverage of simple rules for dataset cases at the time, most typical entity recognition datasets do not include numerical types (e.g., CoNLL 2002/2003).

Later datasets like OntoNotes~\cite{ontonotes5}, recognize the importance of such entities (e.g., Percent, Money, Quantity, Ordinal, and Cardinal). But such inclusion is not commonplace and annotation is still restricted only to tagging/extraction.

%slot tagging, but too little language variation and don't generalize, like ATIS - 1994

As more complex tasks from QA %question answering
to document understand get traction, interest has shined again on numerical entities and their importance, especially in domains such as finance and healthcare. This is evidenced by both new tasks like the FinNum series~\cite{ChenHTC19,chenHTC20,chenHTC21} and works in Health IE~\cite{jagannatha2019overview}.

However, such newer datasets are too domain-specific (e.g., buy price, sell price, and stop loss in FinNum) or target only extraction. Not to mention a lack of consistency between annotation types.

% numER dataset as sub-task. quality too bad, not mention now.
% numer digit-only... and mix date/time

Moreover semantics encoded in numerical entities both capture type information and often denote units of measurements. Considering only the numeric value results in loss of knowledge (e.g., magnitude or sub-type compatibility)~\cite{gesese2021}. Other challenges also include variability in mention forms, un-anchored ordinals, and range expressions. Ideally all of which should be interpretable as they significantly change the semantics of a given mention (e.g., "one"; 
%"last";
"30+"; "half"). 

Alternate works like AMR~\cite{banarescu-etal-2013-abstract} also recognize the importance of representing and tagging numerical and temporal terms (e.g., :quant, :unit, :year, :season, :weekday, etc.), but do not address typing and purposefully do not perform any normalization. Moreover, their 50-pages annotation guidelines are overly complex.

%Relations for quantities. :quant, :unit,
%:scale.
%• Relations for date-entities. :day, :month,
%:year, :weekday, :time, :timezone, :quarter,
%:dayperiod, :season, :decade, :century, :calendar, :era.

\subsection{Temporal Entities}
\label{subsec:temporal entities}

Differently from numerical expressions, there have been frequent efforts in creating
corpora and assessing their quality regarding temporal entities. Such as the TempEval series~\cite{verhagen-etal-2007-semeval,verhagen-etal-2010-semeval,uzzaman-etal-2013-semeval} and annotated corpora like ACE\footnote{LDC2005T07 and LDC2006T06 in the LDC
catalogue. %(http://www.ldc.upenn.edu)
} and TimeBank\footnote{LDC2006T08 in the LDC catalogue.}.

However, even with a certain degree of maturity, most annotation approaches suffer from low coverage of mention forms, too high complexity, difficulty to extend, and lack of granularity for downstream use. The most popular annotation standards are TIDES TIMEX2~\cite{ferro2005tides} and TimeML’s TIMEX3~\cite{james2005}, which are used in datasets like TempEval and WikiWars~\cite{10.5555/1870658.1870747}.
% WikiWars itself has been created to address limitations in existing datasets. 

The limitations of TIMEX2, for example, span the annotation of time zones, event-based expressions, duration and set anchor restrictions.\footnote{Sentences like "every Tuesday since March" or "five days in mid-August" can not be precisely annotated}
Moreover annotation guidelines for such schemas are complex, abstract, and sometimes open to ambiguous interpretation~\cite{Sauri2006}.

Previous attempts to address temporal expression recognition and resolution have also highlighted other limitations of such schemas. From problems in the standard evaluation datasets~\cite{li2014chinese} (with missing and incorrect annotations), to specific cumbersome annotation requirements such as "Empty tags are TIMEX3 tags that do not contain any tokens and should be created whenever a temporal expression can be inferred from preexisting text-consuming TIMEX3 tags" which is either not applied or inconsistently done~\cite{manfredi2014heideltime}, leading to issues with anchored durations (e.g., "a month ago") and range expressions that combine two TIMEX3 tags (e.g., "from 2010 to 2014").

Support for temporal ranges in general is non-intuitive. For instance, expressions like "from 3 to 4 p.m." to "12-13 March 2011" are hard or impossible to annotate for their ambiguities, and when annotated or generated in the response of extraction, are hard to be used downstream.

In order to reduce complexity and cover more mention forms in an easy to consume way, our time expressions differ from TIMEX2/TIMEX3. Mostly in the granularity of types and representation of complex expressions like durations and recurring datetimes.\footnote{E.g., TimeML tags "November" as Date, while NTX tags it as DateRange}

Variety of mention forms highlight also the need for consistency between numerical and temporal expression recognition. Time will depend on numbers. For example "in half an hour" requires consistent handling of fraction term and articles. 

Lastly, many common mention forms like "8:24 a.m. Chicago time" are not well covered by previous guidelines, but are covered in NTX.

% WikiWars TIMEX2 tags 2600 TIMEX2
% English Wikipedia, which we have annotated with TIMEX2 tags \cite{10.5555/1870658.1870747}

%TIMEX2 guidelines stipulate that the anchors of durations cannot have a MOD attribute, so if the anchor is mid-August, the value of the anchor must refer to August

%the tasks of identifying and assigning values to temporal expressions have recently received significant attention, resulting in the creation of mature corpus annotation guidelines (e.g. TIMEX2 and TimeML)

%has established itself as an important and promising research area as it greatly helps the accessibility and usability of existing 

\section{Dataset Details}
\label{sec:dataset details}

To alleviate some of the described issues and trying to cover a wide variety of scenarios
we propose a new dataset - NTX (Numerical and Temporal eXpressions) - for the evaluation of numerical and temporal recognition systems.

NTX was build on real usage over the past several years and targets cross-domain scenarios and the interrelated nature of numex and timex. Coverage includes variants of languages (e.g., French covers both fr-FR and fr-CA) and formal and informal mention forms. 
%conversation scenarios, short documents

The dataset covers 14 languages - English, Chinese, Dutch, French, German, Italian, Japanese, Korean, Portuguese, Spanish, Swedish, Turkish, Hindi, and Arabic; which belong to a diverse set of language families. With 8 sub-types of numerical entities (cardinal, ordinal, percentage, numerical range, age, currency, dimension, and temperature) and 10 temporal sub-types (date, time, datetime, date range, time range, datetime range, holiday, duration, timezone, and recurring set). 
%info in dimension can also be used for sub-types.
%including variants as in the wild you don't know

These entities both provide fine-grained granularity, many times required by downstream tasks, and help address the previously mentioned limitations of other datasets.

This is accomplished mainly in two fronts: i) Allowing fine-grained types that keep semantics useful in downstream tasks (i.e., not mixing the concepts of date, time, and ranges) and adding new sub-types for previously not supported annotations (e.g., such as holidays for mentions like "Xmas", "Easter Sunday", etc.); and ii) Simplifying annotation by, instead of having complex multi-level annotations (entities, relationships, and modifiers annotated in different ways), grouping them as much as possible and representing them in a streamlined entity-level form (i.e., permit combination of modifiers, representing a range by start/end/length, instead of complex relationships). For example, "after mid-August" is 1 entity (of date range type), and not 3 entities plus additional relationship annotations.

%Distrib in table \ref{tab:dataset distribution numex, tab:dataset distribution timex} 

% limitation: co-reference, temporal relations.

% WikiWars TIMEX2 tags 2600 TIMEX2

%pragmatic rules no linguistic requirements
%resource-based, easy for people to customize to extend or to disable what don't want

%data distribution
%rare cases
%english foreign influenced...

The dataset contains over 26000 sentences across the different languages.

To fulfil its requirement of covering multiple scenarios, it includes both long and short sentences. Specifically to cover common conversational scenarios, where lack of context is commonplace, approximately $13\%$ of cases consist of only an entity mention (i.e., they could be the input directly to normalization and resolution as-is).

Moreover, the dataset includes not only sentences with entity mentions, but also multiple sentences that correctly have no annotation. This is necessary to make sure evaluations also  cover behaviour related to false positives. Table \ref{tab:data_dist_summary} shows a summary of sentence counts in the dataset\footnote{Detailed statistics by sub-type are shown in Tables \ref{tab:dataset distribution numex} and \ref{tab:dataset distribution timex} in Appendix \ref{sec:datasetstats}.}.

\begin{figure}[t]
\centering
\includegraphics[width=\linewidth]{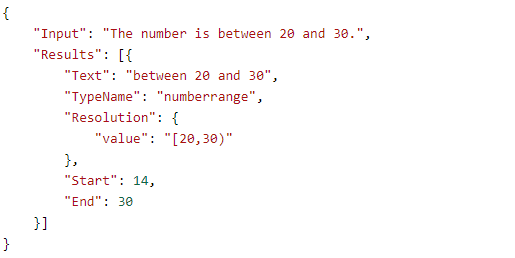}
\caption{
Number range annotation.
}
\label{fig:number range json}
\end{figure}

\begin{figure}[t]
\centering
\includegraphics[width=\linewidth]{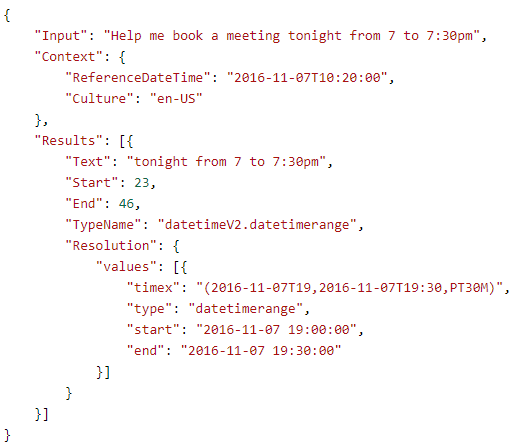}
\caption{
DateTime range annotation.
}
\label{fig:datetime range json}
\end{figure}

Data is made available in the form of JSON files in order to represent not only the tagging of entities, but also a representation of mention normalization and potential resolutions.
Figures \ref{fig:number range json} and \ref{fig:datetime range json} show two example mentions.

\begin{table*}
\resizebox{\textwidth}{!}{
\small\centering
\begin{tabular}{clcccccccccccccc}
%	\toprule
%	& &\multicolumn{14}{c}{Language} \\
%	&   &\rotr{Arabic} & \rotr{Chinese} & \rotr{Dutch} & \rotr{English} & \rotr{French} & \rotr{German} & \rotr{Hindi} & \rotr{Italian} & \rotr{Japanese} & \rotr{Korean} & \rotr{Portuguese}                                        & \rotr{Spanish} & \rotr{Swedish}                                            & \rotr{Turkish} \\ \midrule
&   &AR & ZH & NL & EN & FR & DE & HI & IT & JA & KO & PT & ES & SV & TR \\ \midrule
&w/ numerical entities      & 526                  & 1068                 & 433                  & 1127                 & 696                  & 207                  & 582                  & 436                  & 1442                 & 713                  & 647                  & 691                  & 394                  & 457                 
      \\
&w/ temporal entities        & 909                  & 518                  & 2064                 & 1920                 & 2500                 & 480                  & 1368                 & 808                  & 1496                 & 1004                 & 526                  & 1660                 & 461                  & 962                 
       \\ 
% &Overall &2131                 & 1892                 & 3695                 & 5488                 & 5761                 & 978                  & 2787                 & 1937                 & 3931                 & 2358                 & 1564                 & 3050                 & 703                  & 2198           \\
&Overall &1762                 & 2007                 & 2778                 & 3546                 & 3677                 & 769                  & 2229                 & 1434                 & 3796                 & 2057                 & 1409                 & 2661                 & 1043                 & 1564                 \\
      \bottomrule
\end{tabular}}
\caption{Numbers of sentences with 
%numerical and temporal 
entities by language. \emph{Overall} includes sentences with no entity.}
\label{tab:data_dist_summary}
\end{table*}

%tonight as an example

%\begin{lstlisting}[language=json,firstnumber=1]
%{
% "Input": "The number is between 20 and 30.",
% "Results": [
%  {
%   "Text": "between 20 and 30",
%   "TypeName": "numberrange",
%   "Resolution": {
%    "value": "[20,30)"
%   },
%   "Start": 14,
%   "End": 30
%  }
% ]
%}
%\end{lstlisting}
%
%And a datetime range:
%
%\begin{lstlisting}[language=json,firstnumber=1]
%{
% "Input": "Help me book a meeting tonight from 7 to 7:30pm",
% "Context": {
%  "ReferenceDateTime": "2016-11-07T10:20:00"
% },
% "Results": [
%  {
%   "Text": "tonight from 7 to 7:30pm",
%   "Start": 23,
%   "End": 46,
%   "TypeName": "datetimeV2.datetimerange",
%   "Resolution": {
%    "values": [
%     {
%      "timex": "(2016-11-07T19,2016-11-07T19:30,PT30M)",
%      "type": "datetimerange",
%      "start": "2016-11-07 19:00:00",
%      "end": "2016-11-07 19:30:00"
%     }
%    ]
%   }
%  }
% ]
%}
%\end{lstlisting}  

%Handling of ambiguous phrases - spring by itself
%rules to filter...

Nonetheless, resolution of relative expressions or un-anchored mentions can still be ambiguous.
In such cases the dataset lists both future and past resolutions as acceptable. For example, "Friday" can be interpreted either as "upcoming Friday" or "past Friday".

%\subsection{Dataset creation}
%\label{subsec:dataset creation}

Details of the dataset creation are provided in Appendix \ref{sec:datasetcreation}.

%tools to generate labeled data
%and normalization...

% OntoNotes 5 corpus has
%the types Percent, Money, Quantity, Ordinal, and Cardinal

%NumER: A Fine-Grained Numeral Entity Recognition Dataset
%too unwiedly 
%dataset not there,  2,481 sentences
%---we EN 1278

%date and year as num-only
%quantity units not included in number and only cover 7 "unit"

%musd/informal...

%we also show that Patterns/rules can be made to a large degree reusable between languages in different families.

\section{Rule-Based System Design}
\label{sec:implementation}

While the limitations of rule-based systems are well known, especially in regards to the maintenance of large amounts or rules and their interactions, we have opted in Recognizers-Text~\cite{recognizers-text} for a rule-based design due to three key design principles:

i) determinism: the system needs to always produce predictable output, so downstream consumers of it's output can act accordingly;
ii) prioritize recall: as rules are prone to false positives or false negatives, the system should focus on coverage to the extent possible, as false positives could potentially be filtered in pre- or post-processing stages.
iii) no need for expert knowledge to make changes: rules are a somewhat straightforward way to represent knowledge for entity extraction, instead of requiring users to have a linguistic or machine learning background. 

Although neural architectures have shown to be able to perform competitively on time expression recognition on previous datasets~\cite{lange-etal-2020-adversarial,chen2019exploring,cao-etal-2022-xltime}, such architectures don't address the requirements above, and both their implementation and evaluation target only recognition.

The core structure of the rule-based extractors is inspired by SUTIME~\cite{chang-manning-2012-sutime} and basically follows its \emph{three-types-of-rules} design. Rules are roughly categorized into mention capture, composition, and filtering. The same structured core of rules is shared across languages while localized for language-specific properties.
We utilize regular expressions throughout the system.

For better performance, to avoid unwieldy long regexes, the currency and timezone extractors also make use of dictionaries, in the form of tries (prefix trees) for tagging. This has the additional benefit of facilitating users scenarios where they require to load their own extra terms.

Normalization and resolution (together termed parsing in our system) however, are key, and require additional code writing. Parsing takes the form of a cascade of parsers per type increasing in complexity.
%We design and implement two core flows that are shared across sets of languages (i.e., a CJK core covering Chinese, Japanese, and Korean; and a default core for all others). 

Language-specific behaviour is defined as overridable functions in each language configuration. If a function is not overrided, the core default behaviour is adopted (with other language-specific configurations).

%There are many methods focused on leveraging contextual word representation for tagging multilingual time expressions~\cite{lange-etal-2020-adversarial,chen2019exploring,cao-etal-2022-xltime}
%While neural parsers of time expressions are limited in English and only parsing simple sub-type such as time intervals~\cite{laparra-etal-2018-characters,xu-etal-2019-pre}.

%rule-based HeidelTime

% \section{Analysis}
% \label{sec:analysis}

% Please refer to Appendix \ref{sec:entity retrieval performance} for more 

%\subsection{Case Study}
% \label{subsec:case study}

\section{Conclusion}
\label{sec:conclusion}

Here we propose NTX, a novel multi-lingual evaluation dataset covering diverse temporal and numerical expressions across 14 languages. %and extraction, normalization, and resolution.
We also provide Recognizers-Text as a robust baseline system for comparisons against other models over this evaluation dataset.

\section*{Limitations}
\label{sec:limitations}

As NTX includes not only span detection, but also type information and resolution, potential ambiguities in type are important. Where ambiguities were detected during the dataset creation, consensus of expert annotators was used to determine the most likely (most commonly applicable) type. We plan for future versions of the dataset to include information about such alternative type interpretations.

The described rule-based system (Recognizers-Text) is intended primarily as a strong evaluation baseline over NTX for performance comparisons, but it can also serve as a potential source of automatically labeled data in the NTX schema for training semi-supervised models\footnote{Different implementations of the system are available in the GitHub repository, which may present differing output quality. The .NET version is recommended as the canonical version for evaluations.}. As mentioned in Section \ref{sec:implementation}, rule-bases system have well know limitations, including handling of false positives. Utilizing the described baseline system to automatically generate annotated data must be followed by annotation review to account for such issues.

It is also important to note that, while the dataset contains the data for temporal expressions in SV, KO, and AR, as well as numerical expressions with units in AR, these are currently not supported by the baseline rule-bases system.

Lastly, the current dataset schema may be somewhat unintuitive to manual inspection; as it focuses on extraction/parsing representation. The open-sourced code includes evaluation scripts to calculate Precision/Recall/F-1 over it, for ease of use.

\bibliography{main}

\begin{thebibliography}{24}
\expandafter\ifx\csname natexlab\endcsname\relax\def\natexlab#1{#1}\fi

\bibitem[{Alonso et~al.(2007)Alonso, Gertz, and
  Baeza-Yates}]{10.1145/1328964.1328968}
Omar Alonso, Michael Gertz, and Ricardo Baeza-Yates. 2007.
\newblock \href {https://doi.org/10.1145/1328964.1328968} {On the value of
  temporal information in information retrieval}.
\newblock \emph{SIGIR Forum}, 41(2):35–41.

\bibitem[{Banarescu et~al.(2013)Banarescu, Bonial, Cai, Georgescu, Griffitt,
  Hermjakob, Knight, Koehn, Palmer, and
  Schneider}]{banarescu-etal-2013-abstract}
Laura Banarescu, Claire Bonial, Shu Cai, Madalina Georgescu, Kira Griffitt, Ulf
  Hermjakob, Kevin Knight, Philipp Koehn, Martha Palmer, and Nathan Schneider.
  2013.
\newblock \href {https://aclanthology.org/W13-2322} {{A}bstract {M}eaning
  {R}epresentation for sembanking}.
\newblock In \emph{Proceedings of the 7th Linguistic Annotation Workshop and
  Interoperability with Discourse}, pages 178--186, Sofia, Bulgaria.
  Association for Computational Linguistics.

\bibitem[{Cao et~al.(2022)Cao, Groves, Saha, Tetreault, Jaimes, Peng, and
  Yu}]{cao-etal-2022-xltime}
Yuwei Cao, William Groves, Tanay~Kumar Saha, Joel Tetreault, Alejandro Jaimes,
  Hao Peng, and Philip Yu. 2022.
\newblock \href {https://aclanthology.org/2022.findings-naacl.148} {{XLT}ime: A
  cross-lingual knowledge transfer framework for temporal expression
  extraction}.
\newblock In \emph{Findings of the Association for Computational Linguistics:
  NAACL 2022}, pages 1931--1942, Seattle, United States. Association for
  Computational Linguistics.

\bibitem[{Chang and Manning(2012)}]{chang-manning-2012-sutime}
Angel~X. Chang and Christopher Manning. 2012.
\newblock \href
  {http://www.lrec-conf.org/proceedings/lrec2012/pdf/284_Paper.pdf} {{SUT}ime:
  A library for recognizing and normalizing time expressions}.
\newblock In \emph{Proceedings of the Eighth International Conference on
  Language Resources and Evaluation ({LREC}'12)}, pages 3735--3740, Istanbul,
  Turkey. European Language Resources Association (ELRA).

\bibitem[{Chen et~al.(2019{\natexlab{a}})Chen, Huang, Takamura, and
  Chen}]{ChenHTC19}
Chung{-}Chi Chen, Hen{-}Hsen Huang, Hiroya Takamura, and Hsin{-}Hsi Chen.
  2019{\natexlab{a}}.
\newblock Final report of the {NTCIR-14} finnum task: Challenges and current
  status of fine-grained numeral understanding in financial social media data.
\newblock In \emph{{NII} Testbeds and Community for Information Access Research
  - 14th International Conference, {NTCIR} 2019, Tokyo, Japan, June 10-13,
  2019, Revised Selected Papers}, volume 11966 of \emph{Lecture Notes in
  Computer Science}, pages 183--192. Springer.

\bibitem[{Chen et~al.(2020)Chen, Huang, Takamura, and Chen}]{chenHTC20}
Chung{-}Chi Chen, Hen{-}Hsen Huang, Hiroya Takamura, and Hsin{-}Hsi Chen. 2020.
\newblock Overview of the ntcir-15 finnum-2 task: Numeral attachment in
  financial tweets.
\newblock In \emph{Proceedings of the 15th NTCIR Conference on Evaluation of
  Information Access Technologies}, pages 75--78.

\bibitem[{Chen et~al.(2021)Chen, Huang, Takamura, and Chen}]{chenHTC21}
Chung{-}Chi Chen, Hen{-}Hsen Huang, Hiroya Takamura, and Hsin{-}Hsi Chen. 2021.
\newblock Overview of the ntcir-16 finnum-3 task: Investor's and manager's
  fine-grained claim detection.
\newblock In \emph{Proceedings of the 16th NTCIR Conference on Evaluation of
  Information Access Technologies}.

\bibitem[{Chen et~al.(2019{\natexlab{b}})Chen, Wang, and
  Karlsson}]{chen2019exploring}
Sanxing Chen, Guoxin Wang, and Börje~F. Karlsson. 2019{\natexlab{b}}.
\newblock \href
  {https://www.microsoft.com/en-us/research/publication/exploring-word-representations-on-time-expression-recognition/}
  {Exploring word representations on time expression recognition}.
\newblock Technical Report MSR-TR-2019-46, Microsoft Research.

\bibitem[{Ferro et~al.(2005)Ferro, Gerber, Mani, Sundheim, and
  Wilson}]{ferro2005tides}
Lisa Ferro, Laurie Gerber, Inderjeet Mani, Beth Sundheim, and George Wilson.
  2005.
\newblock Tides: 2005 standard for the annotation of temporal expressions.
\newblock Technical report, MITRE CORP MCLEAN VA.

\bibitem[{Gesese et~al.(2021)Gesese, Biswas, Alam, and Sack}]{gesese2021}
Genet~Asefa Gesese, Russa Biswas, Mehwish Alam, and Harald Sack. 2021.
\newblock \href {https://doi.org/10.3233/SW-200404} {A survey on knowledge
  graph embeddings with literals: Which model links better literal-ly?}
\newblock \emph{Semantic Web}, 12(4):617--647.

\bibitem[{Grudin and Jacques(2019)}]{10.1145/3290605.3300439}
Jonathan Grudin and Richard Jacques. 2019.
\newblock \href {https://doi.org/10.1145/3290605.3300439} {Chatbots, humbots,
  and the quest for artificial general intelligence}.
\newblock In \emph{Proceedings of the 2019 CHI Conference on Human Factors in
  Computing Systems}, CHI '19, page 1–11, New York, NY, USA. Association for
  Computing Machinery.

\bibitem[{Huang et~al.(2017)Huang, Lin, McConnell, and
  Karlsson}]{recognizers-text}
Wenhao Huang, Zijia Lin, Chris McConnell, and B{\"{o}}rje~F. Karlsson. 2017.
\newblock \href {https://doi.org/10.5281/zenodo.6860598} {{Recognizers-Text}:
  {R}ecognition and resolution of numbers, units, and date/time entities
  expressed across multiple languages}.

\bibitem[{Jagannatha et~al.(2019)Jagannatha, Liu, Liu, and
  Yu}]{jagannatha2019overview}
Abhyuday Jagannatha, Feifan Liu, Weisong Liu, and Hong Yu. 2019.
\newblock Overview of the first natural language processing challenge for
  extracting medication, indication, and adverse drug events from electronic
  health record notes (made 1.0).
\newblock \emph{Drug safety}, 42(1):99--111.

\bibitem[{Lange et~al.(2020)Lange, Iurshina, Adel, and
  Str{\"o}tgen}]{lange-etal-2020-adversarial}
Lukas Lange, Anastasiia Iurshina, Heike Adel, and Jannik Str{\"o}tgen. 2020.
\newblock \href {https://doi.org/10.18653/v1/2020.repl4nlp-1.14} {Adversarial
  alignment of multilingual models for extracting temporal expressions from
  text}.
\newblock In \emph{Proceedings of the 5th Workshop on Representation Learning
  for NLP}, pages 103--109, Online. Association for Computational Linguistics.

\bibitem[{Li et~al.(2014)Li, Str{\"o}tgen, Zell, and Gertz}]{li2014chinese}
Hui Li, Jannik Str{\"o}tgen, Julian Zell, and Michael Gertz. 2014.
\newblock Chinese temporal tagging with heideltime.
\newblock In \emph{Proceedings of the 14th Conference of the European Chapter
  of the Association for Computational Linguistics, volume 2: Short Papers},
  pages 133--137.

\bibitem[{Manfredi et~al.(2014)Manfredi, Str{\"o}tgen, Zell, and
  Gertz}]{manfredi2014heideltime}
Giulio Manfredi, Jannik Str{\"o}tgen, Julian Zell, and Michael Gertz. 2014.
\newblock Heideltime at eventi: Tuning italian resources and addressing
  timeml's empty tags.
\newblock \emph{HeidelTime at EVENTI: Tuning Italian Resources and Addressing
  TimeML's Empty Tags}, pages 39--43.

\bibitem[{Mazur and Dale(2010)}]{10.5555/1870658.1870747}
Pawet Mazur and Robert Dale. 2010.
\newblock Wikiwars: A new corpus for research on temporal expressions.
\newblock In \emph{Proceedings of the 2010 Conference on Empirical Methods in
  Natural Language Processing}, EMNLP '10, page 913–922, USA. Association for
  Computational Linguistics.

\bibitem[{Palmer and Day(1997)}]{10.3115/974557.974585}
David~D. Palmer and David~S. Day. 1997.
\newblock \href {https://doi.org/10.3115/974557.974585} {A statistical profile
  of the named entity task}.
\newblock In \emph{Proceedings of the Fifth Conference on Applied Natural
  Language Processing}, ANLC '97, page 190–193, USA. Association for
  Computational Linguistics.

\bibitem[{Pustejovsky et~al.(2005)Pustejovsky, Knippen, Littman, and
  Saurí}]{james2005}
James Pustejovsky, Robert Knippen, Jessica Littman, and Roser Saurí. 2005.
\newblock \href {http://www.jstor.org/stable/30200549} {Temporal and event
  information in natural language text}.
\newblock \emph{Language Resources and Evaluation}, 39(2/3):123--164.

\bibitem[{Saurí et~al.(2006)Saurí, Littman, Gaizauskas, Setzer, and
  Pustejovsky}]{Sauri2006}
Roser Saurí, Jessica Littman, Robert Gaizauskas, Andrea Setzer, and James
  Pustejovsky. 2006.
\newblock Timeml annotation guidelines, version 1.2.1.

\bibitem[{UzZaman et~al.(2013)UzZaman, Llorens, Derczynski, Allen, Verhagen,
  and Pustejovsky}]{uzzaman-etal-2013-semeval}
Naushad UzZaman, Hector Llorens, Leon Derczynski, James Allen, Marc Verhagen,
  and James Pustejovsky. 2013.
\newblock \href {https://aclanthology.org/S13-2001} {{S}em{E}val-2013 task 1:
  {T}emp{E}val-3: Evaluating time expressions, events, and temporal relations}.
\newblock In \emph{Second Joint Conference on Lexical and Computational
  Semantics (*{SEM}), Volume 2: Proceedings of the Seventh International
  Workshop on Semantic Evaluation ({S}em{E}val 2013)}, pages 1--9, Atlanta,
  Georgia, USA. Association for Computational Linguistics.

\bibitem[{Verhagen et~al.(2007)Verhagen, Gaizauskas, Schilder, Hepple, Katz,
  and Pustejovsky}]{verhagen-etal-2007-semeval}
Marc Verhagen, Robert Gaizauskas, Frank Schilder, Mark Hepple, Graham Katz, and
  James Pustejovsky. 2007.
\newblock \href {https://aclanthology.org/S07-1014} {{S}em{E}val-2007 task 15:
  {T}emp{E}val temporal relation identification}.
\newblock In \emph{Proceedings of the Fourth International Workshop on Semantic
  Evaluations ({S}em{E}val-2007)}, pages 75--80, Prague, Czech Republic.
  Association for Computational Linguistics.

\bibitem[{Verhagen et~al.(2010)Verhagen, Saur{\'\i}, Caselli, and
  Pustejovsky}]{verhagen-etal-2010-semeval}
Marc Verhagen, Roser Saur{\'\i}, Tommaso Caselli, and James Pustejovsky. 2010.
\newblock \href {https://aclanthology.org/S10-1010} {{S}em{E}val-2010 task 13:
  {T}emp{E}val-2}.
\newblock In \emph{Proceedings of the 5th International Workshop on Semantic
  Evaluation}, pages 57--62, Uppsala, Sweden. Association for Computational
  Linguistics.

\bibitem[{Weischedel et~al.()Weischedel, Palmer, Marcus, Hovy, Pradhan,
  Ramshaw, Xue, Taylor, Kaufman, Franchini, El-Bachouti, Belvin, and
  Houston}]{ontonotes5}
Ralph Weischedel, Martha Palmer, Mitchell Marcus, Eduard Hovy, Sameer Pradhan,
  Lance Ramshaw, Nianwen Xue, Ann Taylor, Jeff Kaufman, Michelle Franchini,
  Mohammed El-Bachouti, Robert Belvin, and Ann Houston.
\newblock \href {https://doi.org/10.35111/xmhb-2b84} {{OntoNotes Release 5.0}}.
\newblock \emph{LDC2013T19}.

\end{thebibliography}

\clearpage
\newpage
% \onecolumn

\appendix

\begin{table*}[t]
\small\centering
%\resizebox{\textwidth}{!}{
\begin{tabular}{clcccccccccccccc}
\toprule
& &\multicolumn{14}{c}{Language} \\
          &     &  \rotr{Arabic} & \rotr{Chinese} & \rotr{Dutch} & \rotr{English} & \rotr{French} & \rotr{German} & \rotr{Hindi} & \rotr{Italian} & \rotr{Japanese} & \rotr{Korean} & \rotr{Portuguese}                                        & \rotr{Spanish} & \rotr{Swedish}                                            & \rotr{Turkish} \\ \midrule
               \multirow{9}{*}{\rotrow{Sub-type}}
&Cardinal        & 284    & 362     & 162   & 233     & 273    & 55     & 192   & 152     & 764      & 254    & 174                                               & 206     & 152                                                & 129     \\
&Number Range    & 145    & 65      & 53    & 87      & 85     & 6      & 76    & 34      & 185      & 132    &  50 & 98      & 30  & 61      \\
& Ordinal         & 70     & 7       & 48    & 54      & 70     & 38     & 64    & 31      & 101      & 68     & 33                                                & 82      & 43                                                 & 39      \\
&Percentage      & 27     & 152     & 13    & 20      & 35     & 15     & 16    & 11      & 204      & 30     & 51                                                & 66      & 11                                                 & 11      \\
&Age             & 0      & 10      & 17    & 19      & 18     & 14     & 21    & 15      & 18       & 19     & 18                                                & 18      & 20                                                 & 18      \\
&Currency        & 0      & 35      & 32    & 180     & 114    & 39     & 115   & 104     & 68       & 108    & 124                                               & 122     & 36                                                 & 109     \\
&Dimension       & 0      & 36      & 74    & 93      & 60     & 28     & 53    & 55      & 64       & 64     & 65                                                & 54      & 67                                                 & 56      \\
&Temperature     & 0      & 10      & 34    & 36      & 41     & 12     & 45    & 34      & 38       & 38     & 47                                                & 45      & 35                                                 & 34      \\
%&Phone Number    &        & 391     &       & 405     &        &        &       &         &          &        & 85                                                &         &                                                    &         \\\bottomrule
\bottomrule
\end{tabular}%}
\caption{Numbers of sentences with numerical entities by language and sub-type.}
\label{tab:dataset distribution numex}
\end{table*}

\begin{table*}[!t]
%\resizebox{\textwidth}{!}{
\small\centering
\begin{tabular}{clcccccccccccccc}
	\toprule
	& &\multicolumn{14}{c}{Language} \\
	&     &  \rotr{Arabic} & \rotr{Chinese} & \rotr{Dutch} & \rotr{English} & \rotr{French} & \rotr{German} & \rotr{Hindi} & \rotr{Italian} & \rotr{Japanese} & \rotr{Korean} & \rotr{Portuguese}                                        & \rotr{Spanish} & \rotr{Swedish}                                            & \rotr{Turkish} \\ \midrule
	\multirow{9}{*}{\rotrow{Sub-type}}
&Date            & 118    & 64      & 226   & 148     & 228    & 28     & 122   & 107     & 214      & 194    & 65                                                & 73      & 132                                                & 94      \\
&Date Range      & 305    & 67      & 319   & 352     & 514    & 63     & 318   & 241     & 246      & 180    & 60                                                & 374     & 60  & 221     \\
&DateTime        & 72     & 15      & 134   & 81      & 125    & 21     & 67    & 68      & 87       & 59     & 61                                                & 63      & 15  & 64      \\
&DateTime Range  & 73     & 27      & 180   & 96      & 183    & 32     & 86    & 77      & 99       & 62     & 41                                                & 118     & 27  & 67      \\
&Model - Overall & 112    & 257     & 741   & 916     & 884    & 202    & 507   & 114     & 584      & 316    & 156                                               & 825     & 112 & 316     \\
&Duration        & 55     & 15      & 87    & 62      & 103    & 21     & 61    & 44      & 44       & 34     & 21                                                & 23      & 15  & 33      \\
&Holiday         & 18     & 35      & 46    & 26      & 50     & 61     & 30    & 11      & 43       & 15     & 14                                                & 17      & 11  & 19      \\
&Set             & 27     & 8       & 52    & 32      & 58     & 13     & 30    & 27      & 33       & 27     & 20                                                & 18      & 8   & 25      \\
&Time            & 74     & 14      & 138   & 93      & 198    & 26     & 84    & 67      & 79       & 61     & 58                                                & 56      & 14  & 69      \\
&Time Range      & 55     & 16      & 122   & 65      & 109    & 13     & 63    & 52      & 67       & 56     & 30                                                & 93      & 13  & 54      \\
&Timezone        & 0      & 0       & 19    & 49      & 48     & 0      & 0     & 0       & 0        & 0      & 0                                                 & 0       & 54                                                 & 0       \\ \bottomrule
\end{tabular}%}
\caption{Numbers of sentences with temporal entities by language and sub-type.}
\label{tab:dataset distribution timex}
\end{table*}

\section{Dataset creation process}
\label{sec:datasetcreation}

The dataset was created firstly in English and Chinese by using hired specialist vendors with linguistic expertise to generate sentences and utterances covering formal and informal cases in both document-type sentences, as well as, conversational/social media-type utterances. 
The same vendors (level 1 annotators) were tasked with annotating all entity types in the collected data. Upon completion of this stage, two expert annotators (level 2 annotators) validated all annotations and documented existing disagreements and desired behaviour on possibly ambiguous cases.

The baseline rule-based system was developed in parallel and was used to further validate annotation. Any failure cases of the system execution resulted in either corrections in the annotations or in extensions to the system to properly address them. Cases where the annotations are agreed to be correct, but that cannot yet be supported by the system are marked as such in metadata to not break a system run during development (but still included in the dataset).

Language expansion then happened through a mixed process of i) generating data in a similar fashion to English and Chinese, and ii) translating large numbers of examples from the existing languages into new target languages (creating parallel texts). Translation was performed by native speaker vendors. Such process had the benefit of emphasizing a balance between language specific mention formats in generation, while having a certain common coverage across languages via translation.

Moreover, the datasets in each language grew organically through long time usage of the system in a commercial setting, collecting failure feedback cases along with support requests for new scenarios (both added as new examples on the dataset).

%38814 spec input lines

%TimeML complexity and issues

%not only formal docs,
%posters, receipts, forms, documents (id, passport), tickets

%These locales belong to a diverse set of language families- Indo-European, Sino-Tibetan, %Japonic, Semitic, Koreanic , and Turkic.

% \usepackage{booktabs}
% \usepackage{multirow}
\begin{table*}[]\small\centering
\begin{tabular}{@{}cccccccccccc@{}}\toprule
\multicolumn{1}{l}{}       & \multicolumn{1}{l}{} & \multicolumn{1}{c}{Arabic} & \multicolumn{1}{c}{Chinese} & \multicolumn{1}{c}{Dutch} & \multicolumn{1}{c}{English} & \multicolumn{1}{c}{French} & \multicolumn{1}{c}{German} & \multicolumn{1}{c}{Hindi} \\\midrule
\multirow{4}{*}{Sentences} & Distinct & 1380  & 1141  & 3014  & 3133  & 4410  & 809   & 2228  \\
&Avg. Length     & 25.83 & 14.24 & 38.04 & 39.40 & 42.66 & 38.31 & 39.89 \\
%&Med. Length     & 22.00 & 12.00 & 33.00 & 33.00 & 35.00 & 34.00 & 33.00 \\
&Stdev     & 14.22 & 10.00 & 25.71 & 30.05 & 35.24 & 33.61 & 31.11  \\\midrule
\multirow{4}{*}{Entities}  & Total          & 1866                       & 1577                        & 3614                      & 5254                        & 5693                       & 949                        & 2626                      \\
                           & Average           & 1.03                       & 1.07                        & 1.06                      & 1.11                        & 1.09                       & 1.06                       & 1.05                      \\
%                           & Med. Counts          & 1.00                       & 1.00                        & 1.00                      & 1.00                        & 1.00                       & 1.00                       & 1.00                      \\
                           & Stdev          & 0.19                       & 0.32                        & 0.29                      & 0.40                        & 0.36                       & 0.34                       & 0.25                     \\
\toprule
\multicolumn{1}{l}{}       & \multicolumn{1}{l}{} & \multicolumn{1}{c}{Italian} & \multicolumn{1}{c}{Japanese} & \multicolumn{1}{c}{Korean} & \multicolumn{1}{c}{Portuguese} & \multicolumn{1}{c}{Spanish} & \multicolumn{1}{c}{Swedish} & \multicolumn{1}{c}{Turkish} \\\midrule
\multirow{4}{*}{Sentences} & Distinct & 1424  & 2546  & 1747  & 1240  & 2509  & 425   & 1678  \\
&Avg. Length     & 40.89 & 17.10 & 23.89 & 41.67 & 43.00 & 42.40 & 37.61 \\
%&Med. Length     & 29.00 & 15.00 & 20.00 & 30.00 & 35.00 & 32.00 & 30.00 \\
&Stdev     & 39.79 & 12.50 & 17.51 & 41.58 & 35.31 & 39.18 & 33.86                        \\\midrule
\multirow{4}{*}{Entities}  & Total          & 1837                        & 3275                         & 2112                       & 1386                           & 2903                        & 549                         & 2159                        \\
                           & Average           & 1.05                        & 1.07                         & 1.05                       & 1.04                           & 1.06                        & 1.07                        & 1.05                        \\
 %                          & Med. Counts          & 1.00                        & 1.00                         & 1.00                       & 1.00                           & 1.00                        & 1.00                        & 1.00                        \\
                           & Stdev          & 0.25                        & 0.30                         & 0.24                       & 0.28                           & 0.30                        & 0.29                        & 0.26                       
  \\\bottomrule
\end{tabular}
\caption{Statistics of sentence length and entity appearances per language.}
\label{tab:dataset_stats_dist_length_appearances}
\end{table*}

%\begin{table*}[]\small\centering
%\begin{tabular}{@{}cccccccccccc@{}}\toprule
%\multicolumn{1}{l}{}       & \multicolumn{1}{l}{} & \multicolumn{1}{c}{Arabic} & %\multicolumn{1}{c}{Chinese} & \multicolumn{1}{c}{Dutch} & \multicolumn{1}{c}{English} & %\multicolumn{1}{c}{French} & \multicolumn{1}{c}{German} & \multicolumn{1}{c}{Hindi} \\\midrule
%\multirow{4}{*}{Sentences} & Total Counts & 9295 & 20905 & 25419 & 37367 & 41330 & 6130 & 22071 \\
%&Avg. Counts  & 5.15 & 14.24 & 7.45  & 7.87  & 7.88  & 6.86 & 8.80  \\
%&Med. Counts  & 4.00 & 12.00 & 7.00  & 7.00  & 7.00  & 6.00 & 7.00  \\
%&Std. Counts  & 2.87 & 10.00 & 4.27  & 5.68  & 5.89  & 5.18 & 6.47   \\
%\toprule
%\multicolumn{1}{l}{}       & \multicolumn{1}{l}{} & \multicolumn{1}{c}{Italian} & %\multicolumn{1}{c}{Japanese} & \multicolumn{1}{c}{Korean} & \multicolumn{1}{c}{Portuguese} & %\multicolumn{1}{c}{Spanish} & \multicolumn{1}{c}{Swedish} & \multicolumn{1}{c}{Turkish} %\\\midrule
%\multirow{4}{*}{Sentences} & Total Counts & 12818 & 52527 & 13235 & 10442 & 22840 & 3961 & 11886 \\
%&Avg. Counts  & 7.34  & 17.10 & 6.67  & 7.86  & 8.35  & 7.71 & 5.79  \\
%&Med. Counts  & 5.00  & 15.00 & 6.00  & 6.00  & 7.00  & 6.00 & 5.00  \\
%&Std. Counts  & 7.18  & 12.50 & 4.41  & 7.07  & 6.07  & 6.96 & 4.78                         %\\\bottomrule
%\end{tabular}
%\caption{Statistics of sentence length in word counts.}
%\label{tab:sent_length_in_words}
%\end{table*}

\begin{table*}
\resizebox{\textwidth}{!}{
\small\centering
\begin{tabular}{clcccccccccccccc}
	\toprule
	& &\multicolumn{14}{c}{Language} \\
	&   &\rotr{Arabic} & \rotr{Chinese} & \rotr{Dutch} & \rotr{English} & \rotr{French} & \rotr{German} & \rotr{Hindi} & \rotr{Italian} & \rotr{Japanese} & \rotr{Korean} & \rotr{Portuguese}                                        & \rotr{Spanish} & \rotr{Swedish}                                            & \rotr{Turkish} \\ \midrule
\multirow{3}{*}{Entities} & \multicolumn{1}{c}{Timex Counts} & 1618   & 843     & 3307  & 4279    & 5166   & 780    & 2197  & 1539    & 2502     & 1639   & 854        & 2393    & 326     & 1799    \\
                          & Numex Counts                     & 248    & 734     & 307   & 975     & 527    & 169    & 429   & 298     & 773      & 473    & 532        & 510     & 223     & 360     \\
                          & \multicolumn{1}{c}{Total Counts} & 1866   & 1577    & 3614  & 5254    & 5693   & 949    & 2626  & 1837    & 3275     & 2112   & 1386       & 2903    & 549     & 2159   
   \\     
      \bottomrule
\end{tabular}}
\caption{Distribution of numerical and temporal entities.}
\label{tab:ent_dist_summary}
\end{table*}

\begin{table*}
\small\centering
\begin{tabular}{lccc}\toprule
\multicolumn{1}{l}{}  & \# Sentences    & \# Timex Entities    & \# Numex Entities    \\\midrule
TempEval-3 (train)    &  3987               & 1822    & 0             \\
TempEval-3 (test)     &  273               & 138      & 0              \\
Tweets (train)        &   1662              & 892     & 0             \\
Tweets (test)         &   422              & 237     & 0             \\
Wikiwars (train)      &   3822              & 2278    & 0             \\
Wikiwars (test)       &   1537              & 373     & 0              \\
OntoNotes 5.0 (train) &   59924              & 12155   & 13861              \\
OntoNotes 5.0 (dev)   &   8528              & 1721    & 1721              \\
OntoNotes 5.0 (test)  &   8262              & 1814    & 1898           \\\midrule
NTX (English)  & 3133                & 4279    & 975           \\\bottomrule
\end{tabular}
\caption{Overall statistics of different datasets (English).}
\label{tab:stats_comparison_summary}
\end{table*}

\begin{table*}
\small\centering
\begin{tabular}{lrrrrrrrrrr}\toprule
\multicolumn{1}{l}{}  & \multicolumn{1}{l}{} & \multicolumn{4}{c}{\# Timex Entities} & \multicolumn{5}{c}{\# Numex Entities}           \\
\cmidrule(r{4pt}){3-6} \cmidrule(l){7-11}
\multicolumn{1}{l}{}  & Avg. Length    & Date    & Set   & Duration   & Time   & Cardinal & Money & Ordinal & Percent & Quantity \\\midrule
TempEval-3 (train)    & 21.55                & 1505    & 30    & 257        & 30     & -        & -     & -       & -       & -        \\
TempEval-3 (test)     & 22.61                & 96      & 4     & 34         & 4      & -        & -     & -       & -       & -        \\
Tweets (train)        & 7.38                 & 554     & 32    & 167        & 139    & -        & -     & -       & -       & -        \\
Tweets (test)         & 8.01                 & 164     & 6     & 33         & 34     & -        & -     & -       & -       & -        \\
Wikiwars (train)      & 22.51                & 1992    & 19    & 200        & 67     & -        & -     & -       & -       & -        \\
Wikiwars (test)       & 21.66                & 330     & 4     & 22         & 17     & -        & -     & -       & -       & -        \\
OntoNotes 5.0 (train) & 18.16                & 10922   & -     & -          & 1233   & 7367     & 2434  & 1640    & 1763    & 657      \\
OntoNotes 5.0 (dev)   & 17.32                & 1507    & -     & -          & 214    & 938      & 274   & 232     & 177     & 100      \\
OntoNotes 5.0 (test)  & 18.49                & 1602    & -     & -          & 212    & 935      & 314   & 195     & 349     & 105   \\\bottomrule
\end{tabular}
\caption{Detailed statistics of different datasets (English).}
\label{tab:stats_comparison_full}
\end{table*}

\section{Dataset statistics}
\label{sec:datasetstats}
Here we provide detailed sets of statistics on the current dataset version.

The distribution of sub-types per language is shown in Tables \ref{tab:dataset distribution numex} and \ref{tab:dataset distribution timex}. Table \ref{tab:dataset_stats_dist_length_appearances} shows statistics on sentence length and amount of annotated entities per sentence across languages.

Moreover, Table \ref{tab:ent_dist_summary} shows the overall distribution of numerical and temporal entities in the dataset. While Tables \ref{tab:stats_comparison_summary} and \ref{tab:stats_comparison_full} show a comparison of NTX to other common datasets with numeric and temporal expressions and their respective testing splits.

\end{document}